\documentclass[11pt]{article}
\PassOptionsToPackage{table}{xcolor}
\usepackage{acl}
\usepackage{times}
\usepackage[T1]{fontenc}
\usepackage{inconsolata}
\usepackage{graphicx}
\usepackage{amsmath}
\usepackage{booktabs}
\usepackage{float}

\usepackage{latexml}

\iflatexml
  \usepackage{listings}
  \lstnewenvironment{promptbox}{\lstset{basicstyle=\ttfamily\small,columns=fullflexible,breaklines=true}}{}
  \renewcommand{\resizebox}[3]{#3}
\else
  \PassOptionsToPackage{table}{xcolor}
  \usepackage{microtype}
  \usepackage{xfp} 
  \usepackage[most]{tcolorbox}
  
  \newtcblisting{promptbox}{
      enhanced,
      breakable,
      colback=gray!5!white,
      colframe=gray!50!black,
      boxrule=0.5pt,
      arc=2mm,
      left=4pt, right=4pt, top=4pt, bottom=4pt,
      listing only,
      listing options={
          basicstyle=\small\ttfamily,
          breaklines=true,
          columns=fullflexible
      }
  }
\fi

\title{When Does Intrinsic Self-Correction Help? A Task-Sensitive Analysis}

\iflatexml
  \author{
    Elroy Stav\textsuperscript{1,*}, 
    Dvir Berlowitz\textsuperscript{1,*}, 
    Maayan Orner\textsuperscript{2}, 
    Sarit Kraus\textsuperscript{1} \\
    \textsuperscript{1}Department of Computer Science and Artificial Intelligence, Bar-Ilan University \\
    \textsuperscript{2}Independent Researcher \\
    \texttt{\{elroistav, dvir.berl, maayanorner\}@gmail.com}, 
    \texttt{sarit@cs.biu.ac.il} \\
    \small * Equal contribution.
  }
\else
  \author{
    Elroy Stav\textsuperscript{1}\thanks{\ \ Equal contribution.} \quad
    Dvir Berlowitz\textsuperscript{1}\footnotemark[1] \quad
    Maayan Orner\textsuperscript{2} \quad
    Sarit Kraus\textsuperscript{1} \\
    \textsuperscript{1}Department of Computer Science and Artificial Intelligence, Bar-Ilan University \\
    \textsuperscript{2}Independent Researcher \\
    \texttt{\{elroistav, dvir.berl, maayanorner\}@gmail.com} \\
    \texttt{sarit@cs.biu.ac.il}
  }
\fi

\begin{document}

\iflatexml
\else
  \maketitle
\fi

\begin{abstract}
Intrinsic self-correction (SC) aims to improve large language model outputs by prompting a model to revisit its own initial answer without external feedback. Recent studies have questioned the reliability of this approach, showing that models often struggle to judge whether their initial responses are correct. In this work, we take a task-sensitive view of SC. Rather than asking whether it works in general, we examine settings where SC may operate through different mechanisms: verifying explicit constraints, revisiting a complex reasoning process, or providing a second opinion over competing strategies in word-game tasks. Across multiple benchmarks and models, we find that SC can yield consistent performance gains when the underlying task structure facilitates these modes of revision. These results suggest that SC is best understood as a task-dependent inference-time strategy whose usefulness depends on the role the revision stage can play in a given task, rather than as a uniformly reliable method for improving initial model outputs.
\end{abstract}

\section{Introduction}
Large language models (LLMs) often benefit from additional inference-time computation after producing an initial answer. One prominent approach is self-correction, in which a model revisits its own answer, critiques or verifies it, and generates a revised response \citep{Bai2022ConstitutionalAH, Madaan2023SelfRefineIR, Welleck2022GeneratingSB}. This approach is appealing because a model's first answer may not be its best, and a second pass may help recover from initial mistakes.

A central distinction in the self-correction literature concerns the source of feedback used during revision. In some settings, models rely only on their internal knowledge and reasoning to evaluate and revise their outputs. In others, revision is guided by external signals, such as retrieval, search, execution feedback, humans, or other models \citep{Pan2023AutomaticallyCL, Gou2023CRITICLL, Paul2023REFINERRF}. This distinction is important because improvements obtained with external feedback do not necessarily show that models can independently recognize and correct their own mistakes. Intrinsic self-correction (SC), where revision occurs without external feedback, therefore provides a more direct test of whether models can improve their initial responses using only their own internal capabilities.

Prior work shows mixed evidence on the effectiveness of SC. Methods such as Self-Refine and Chain-of-Verification suggest that self-generated critique can improve answers in some cases \citep{Madaan2023SelfRefineIR, Dhuliawala2023ChainofVerificationRH}. However, recent studies argue that these gains are limited or unstable, and that some reported improvements depend on oracle labels or ground-truth feedback rather than purely intrinsic revision. \citet{Huang2023LargeLM} show that LLMs frequently fail to improve on reasoning tasks when asked to self-correct without such feedback, while \citet{Kamoi2024WhenCL} argue that the literature often overstates SC due to impractical frameworks or unfair evaluations. More recent work further highlights risks of negative revision, including answer wavering, prompt bias, cognitive-bias-like effects, and trade-offs between preserving correct answers and correcting incorrect ones \citep{Zhang2024UnderstandingTD, Yang2024ConfidenceVC}. However, other work suggests a more nuanced view: SC may help under particular elicitation and evaluation conditions, e.g., when conditioned on model confidence \citep{Li2024ConfidenceMR, Liu2024LargeLM}.

This paper takes a task-sensitive view of SC. Rather than
asking whether it works generally, we argue that self-correction may work
differently across tasks. We examine three types of tasks that reflect different
possible mechanisms of intrinsic revision. First, we consider verifiable tasks,
where models can assess whether answers satisfy explicit constraints or consistency
requirements. Second, we consider reasoning-intensive tasks, where
a second pass may help the model revisit its reasoning trajectory and produce a
more deliberate solution. Third, we consider word-game tasks, where revision may function less
as binary error detection and more as an intrinsic second opinion over alternative
answers and strategies. We study these in a standard intrinsic setting where the
same model generates, critiques, and revises answers without external tools, oracle
labels, or human feedback.

\emph{Our contribution} is a task-level analysis of intrinsic self-correction. Specifically, we show that its effectiveness depends on task-specific factors and identify settings where it appears to be more promising.

\section{Methodology}

\subsection{Tasks}
{\bf Motivation}
Recent work has shown that SC is limited by the model's inability to reliably judge its own responses \cite{Huang2023LargeLM,Kamoi2024WhenCL}. We select tasks where this limitation may be less prohibitive, or where an additional reasoning pass may still provide value despite it. The tasks we study exemplify three distinct ways in which intrinsic revision may improve performance.

First, in \emph{verifiable} tasks, candidate answers can be evaluated
against explicit constraints, consistency requirements, or task-specific criteria. For example, in SAT, the model can check whether a candidate answer (assignment) satisfies a CNF formula. In such settings, self-correction
may support verification-based repair, where the second pass is used to verify a solution and revise when needed.

Second, in \emph{reasoning-intensive} tasks, self-correction may help by prompting the
model to revisit the reasoning process behind its initial answer and produce a more
deliberate solution.

Third, in our \emph{word-game tasks}, self-correction may function as strategic reconsideration. 
Since these tasks often allow multiple plausible moves rather than a single definitive answer, 
the second pass may be used to compare the initial answer against alternatives, reconsider 
the underlying strategy, and decide whether refinement is warranted. We view this critique 
step as a form of intrinsic advising or second-opinion generation, rather than explicit error detection.

{\bf Benchmarks}
We evaluate the effectiveness of SC on six benchmarks spanning several task settings, including \emph{verifiable} tasks, \emph{reasoning} tasks, and \emph{word games}. 

To examine SC in verifiable and/or reasoning-intensive tasks, we employ three benchmarks: \textbf{SAT}, an NP-complete problem representing extreme verifiability, where finding a satisfying assignment for a Boolean formula in CNF is computationally hard but verification is straightforward; \textbf{BBEH} \cite{kazemi2025BIGBenchEH}, a reasoning-intensive benchmark with many tasks that allow answers to be checked against explicit or reconstructable criteria, making it relevant to both reasoning-based refinement and verification-based repair (we analyze multiple-choice and open-ended subsets separately); and \textbf{HLE} (Humanity's Last Exam) \cite{Phan2025ABO}, for which we use the text-only multiple-choice subset as an expert-level knowledge and reasoning benchmark.

For \textit{word-game tasks}, we adapt three custom environments from the TextArena framework \cite{Guertler2025TextArena} to capture different forms of strategic planning and alternative exploration: \textbf{Wordle}, iterative guess refinement under structured feedback; \textbf{Hangman}, probabilistic strategy revision over multiple turns; and \textbf{Codenames}, multi-role reasoning in which the model acts as both spymaster and guesser. In Codenames, the model generates a single-word clue intended to lead the guesser toward target words while avoiding distractors, testing its ability to reason about word associations, recipient interpretation, and strategic revision.

\subsection{Prompts}
Following \citet{Welleck2022GeneratingSB}, \citet{Saunders2022SelfcritiquingMF}, and \citet{Yang2025TheLO}, we use a two-step self-correction protocol for SAT and word-game benchmarks: (1) prompt the model to produce an initial answer, which also serves as the baseline; (2) prompt the model to revisit the question and its previous answer and produce a revised response. For BBEH and HLE, which involve more complex open-ended reasoning, we follow \citet{Huang2023LargeLM}, \citet{Kim2023LanguageMC}, \citet{Shinn2023ReflexionLA}, and \citet{Bai2022ConstitutionalAH} in using a three-step protocol: (1) generate an initial answer; (2) generate an explicit critique of that answer; (3) produce a revised answer using the critique as guidance. The three-step protocol is better suited to settings where an intermediate critique stage can help surface specific reasoning gaps before revision. Since our analysis compares Base vs. SC within each benchmark rather than across protocols, this difference does not confound our main conclusions. We provide the full prompts used in each benchmark and for each agent in Section~\ref{sec:prompt_templates}.

\section{Experiment}

{\bf Models}
To ensure the robustness and generalizability of our findings, we evaluate the effectiveness of intrinsic self-correction across a diverse suite of five large language models (LLMs). Our selection intentionally spans a spectrum of accessibility paradigms (open-weights versus proprietary APIs), architectural designs, and capability levels.

{\bf Inference and Hyperparameters}
To prioritize deterministic outputs and improve the reproducibility of our reasoning trajectories, we set the sampling temperature to 0 on all benchmarks.

{\bf Evaluation Metrics}
We use \textit{accuracy} (success rate) as our primary evaluation metric for the majority of the benchmarks (SAT, BBEH, HLE).
For the words games, we used task-specific scoring. Wordle uses a turn-based score decreasing linearly from $1$ (first-turn success) to $0$ (failed in all 6 turns); Hangman scores $\frac{M - w}{M}$ on a win (where $M$ is the wrong-guess limit and $w$ wrong guesses were used) and $0$ on a loss; Codenames assigns $+1$ per target word and $-1$ per incorrect selection.
We assess statistical significance ($p < 0.05$) per benchmark using McNemar's test for accuracy tasks, and paired $t$-tests alongside Wilcoxon signed-rank tests for word games.

\section{Results and Discussion}
\label{sec:results}
\begin{table*}[t]
\centering
\resizebox{\linewidth}{!}{
\begin{tabular}{l c c c c c c c c c c}
\toprule
& \multicolumn{2}{c}{\textbf{Mistral Large 2512}} & \multicolumn{2}{c}{\textbf{Claude Haiku 4.5}} & \multicolumn{2}{c}{\textbf{DeepSeek V3.1}} & \multicolumn{2}{c}{\textbf{Gemini 3.1 Flash Lite}} & \multicolumn{2}{c}{\textbf{Gemini Pro}} \\
\cmidrule(lr){2-3} \cmidrule(lr){4-5} \cmidrule(lr){6-7} \cmidrule(lr){8-9} \cmidrule(lr){10-11}
\textbf{Benchmark} & \textbf{Base} & \textbf{SC} & \textbf{Base} & \textbf{SC} & \textbf{Base} & \textbf{SC} & \textbf{Base} & \textbf{SC} & \textbf{Base} & \textbf{SC} \\
\midrule
SAT & 34.62 & \textbf{64.96}$^{*}$ & 75.64 & \textbf{93.59}$^{*}$ & 78.63 & \textbf{92.74}$^{*}$ & 60.68 & \textbf{82.91}$^{*}$ & 97.00 & \textbf{100}$^{*}$ \\
BBEH Non-Choice & 17.04 & \textbf{25.67}$^{*}$ & 28.75 & \textbf{31.92}$^{*}$ & 35.34 & \textbf{46.99}$^{*}$ & 38.20 & \textbf{43.40}$^{*}$ & 80.88 & \textbf{82.30}$^{*}$ \\
BBEH Choice & \textbf{27.75} & 26.50 & \textbf{35.00} & 34.50 & 39.00 & \textbf{42.75}$^{*}$ & 42.00 & \textbf{47.75}$^{*}$ & 84.00 & \textbf{85.07} \\
HLE Text Choice & 9.50 & \textbf{12.50} & 11.00 & \textbf{13.50} & 13.50 & \textbf{20.00}$^{*}$ & 16.50 & \textbf{23.50}$^{*}$ & 34.16 & \textbf{35.40} \\
\midrule
Wordle & \textbf{0.018} & 0.003 & 0.307 & \textbf{0.365}$^{*}$ & 0.039 & \textbf{0.060} & 0.088 & \textbf{0.275}$^{*}$ & 0.423 & \textbf{0.437} \\
Hangman & 34.17 & \textbf{35.83} & 42.92 & \textbf{43.75} & 42.50 & \textbf{51.67}$^{*}$ & 46.67 & \textbf{48.75} & 55.83 & \textbf{58.33} \\
Codenames & 1.178 & \textbf{1.215} & \textbf{0.990} & 0.889 & 0.950 & \textbf{1.030} & 1.198 & \textbf{1.289}$^{*}$ & 1.890 & \textbf{2.068}$^{*}$ \\
\bottomrule
\end{tabular}
}
\caption{Model performance across various benchmarks comparing baseline (Base) and self-correction (SC) settings. Values denote accuracy (\%), except for Wordle and Codenames which employ task-specific scoring metrics. The Gemini Pro column reports results using Gemini 3 Pro Preview for SAT, BBEH, and Codenames, and Gemini 2.5 Pro for HLE, Hangman, and Wordle. $^{*}$ indicates a significant SC improvement over the Baseline ($p < 0.05$).}
\label{tab:model_self_correction}
\end{table*}

\begin{table*}[t]
\centering
\resizebox{\linewidth}{!}{
\begin{tabular}{l c c c c c c c c c c c c c c c c c c c c}
\toprule
& \multicolumn{4}{c}{\textbf{Mistral Large 2512}} & \multicolumn{4}{c}{\textbf{Claude Haiku 4.5}} & \multicolumn{4}{c}{\textbf{DeepSeek V3.1}} & \multicolumn{4}{c}{\textbf{Gemini 3.1 Flash Lite}} & \multicolumn{4}{c}{\textbf{Gemini Pro}} \\
\cmidrule(lr){2-5} \cmidrule(lr){6-9} \cmidrule(lr){10-13} \cmidrule(lr){14-17} \cmidrule(lr){18-21}
\textbf{Benchmark} & \textbf{W2C} & \textbf{C2W} & \textbf{C2C} & \textbf{W2W} & \textbf{W2C} & \textbf{C2W} & \textbf{C2C} & \textbf{W2W} & \textbf{W2C} & \textbf{C2W} & \textbf{C2C} & \textbf{W2W} & \textbf{W2C} & \textbf{C2W} & \textbf{C2C} & \textbf{W2W} & \textbf{W2C} & \textbf{C2W} & \textbf{C2C} & \textbf{W2W} \\
\midrule
SAT   & 32.50 & 2.10 & 32.50 & 32.90 & 17.90 & 0.00 & 75.60 & 6.40 & 14.50 & 0.40 & 78.20 & 6.80 & 22.60 & 0.40 & 60.30 & 16.70 & 3.00 & 0.00 & 97.00 & 0.00 \\
BBEH Non Choice   & 12.70 & 4.10 & 12.90 & 70.20 & 9.90 & 6.80 & 22.00 & 61.30 & 14.10 & 2.40 & 32.90 & 50.60 & 14.80 & 9.60 & 28.60 & 47.00 & 1.80 & 0.40 & 80.50 & 17.30 \\
BBEH Choice     & 8.50 & 9.75 & 18.00 & 63.75 & 12.75 & 13.25 & 21.75 & 52.25 & 7.00 & 3.25 & 35.75 & 54.00 & 15.50 & 9.75 & 32.25 & 42.50 & 4.00 & 2.93 & 81.07 & 12.00 \\
HLE Text Choice & 9.00 & 6.00 & 3.50 & 81.50 & 11.00 & 8.50 & 2.50 & 78.00 & 15.00 & 8.50 & 5.00 & 71.50 & 15.50 & 8.50 & 8.00 & 68.00 & 10.56 & 9.32 & 24.84 & 55.28 \\
\bottomrule
\end{tabular}
}
\caption{Breakdown of the four possible self-correction (SC) outcomes across models and benchmarks. \textbf{W2C} (Improvement: wrong $\rightarrow$ correct), \textbf{C2W} (Regression: correct $\rightarrow$ wrong), \textbf{C2C} (Both Correct: correct $\rightarrow$ correct), and \textbf{W2W} (Both Incorrect: wrong $\rightarrow$ wrong) are expressed as absolute percentages of the total evaluation set.}
\label{tab:sc_dynamics_absolute}
\end{table*}

{\bf General Effectiveness of Self-Correction} 
Pooled across models, SC significantly outperforms the Baseline on all benchmarks. As shown in Table \ref{tab:model_self_correction}, intrinsic self-correction (SC) generally improves performance over zero-shot baselines across the majority of models and task categories. Notably, the SAT tasks show large and consistent improvements under SC, where the mechanism eliminates an average of 65.66\% of the baseline errors. DeepSeek V3.1 and Gemini 3.1 Flash Lite also show consistent, meaningful improvements, achieving average relative performance gains of 26.94\% and 47.27\%, respectively, across all benchmarks. Gemini Pro also improves consistently across all benchmarks, although its gains are smaller, which may reflect the limited remaining room for improvement given its strong baseline performance. These results suggest that, given appropriate task structures, LLMs can successfully refine their initial outputs without external feedback.

Table~\ref{tab:sc_dynamics_absolute} shows that models exhibit qualitatively different SC dynamics on SAT compared to the other benchmarks. On SAT, SC degrades correct answers (C2W) at a rate of at most 2.1\% across all models, while on BBEH and HLE, C2W reaches up to 13.25\%. Although this observation is limited to a single benchmark, the fact that SAT solutions are straightforwardly \emph{verifiable} may explain its distinctively low C2W rate.

{\bf Effectiveness Across Models}
The consistent but smaller gains of Gemini Pro may suggest a diminishing-returns effect for stronger models. Gemini Pro improves only modestly on BBEH and HLE compared to weaker models, indicating that when a model already produces strong initial answers or implicitly self-corrects during reasoning, the revision stage may have less additional value. 

At the other end of the spectrum, very low baseline performance can also limit SC. For example, Mistral performs poorly on Wordle, and SC does not provide meaningful improvement, suggesting that intrinsic revision still depends on the model having sufficient underlying task capability.

{\bf Model-Specific Over-Revision in Codenames}
Claude Haiku 4.5 exhibits performance degradation under SC in the Codenames benchmark. 
Trajectory analysis shows that the model alters its initial clue in approximately 99\% of turns, 
suggesting that, for this model, the revision step almost always replaces the initial strategy rather 
than selectively preserving viable clue-target associations. This pattern is consistent with the trade-off 
described by \citet{Yang2024ConfidenceVC}, in which a stronger tendency to revise can come at the 
cost of preserving initially correct or viable responses.

More broadly, this case highlights a potential 
risk in tasks without a clear notion of correctness: the revision step may encourage unnecessary 
replacement rather than selective improvement. However, since this pattern does not recur across 
other models or across other tasks for the same model, it appears to be a model--task-specific failure mode 
rather than a general limitation of SC. Such unusually high revision rates may therefore serve as 
a diagnostic signal that the revision prompt or task formulation should be adjusted.

{\bf Sub-task-Level Analysis of BBEH}
Although BBEH was selected because it aligns with our reasoning/verifiability hypothesis at the benchmark level, its sub-tasks vary in how strongly they exhibit these properties. We therefore examine whether this within-benchmark variation helps explain where self-correction is most useful. To study this, we use LLM-judges (GPT-5.5 and Opus 4.6) to score each sub-task on reasoning intensity and verifiability. 

Both dimensions are significantly correlated with \emph{the average} accuracy gain from SC across models (reasoning: Spearman's $\rho=0.54$, $p=0.007$; verifiability: $\rho=0.65$, $p<0.001$), as well as for roughly half the models individually.
Notably, models tend to show negligible or negative SC accuracy gains on sub-tasks that score low on both dimensions, like \emph{Movie Recommendation}. 
We provide the scoring setup, sub-task results and judge scores, and additional analyses in Appendix~\ref{sec:per_task_analysis}.

\section{Related Work}

Intrinsic self-correction requires a model to revisit and revise its answers without external supervision. Methods range from direct two-step revision to explicit critique-then-revise pipelines \cite{Saunders2022SelfcritiquingMF,Welleck2022GeneratingSB,Huang2023LargeLM,Kim2023LanguageMC}, with Self-Refine extending this iteratively \cite{Madaan2023SelfRefineIR} and Chain-of-Verification replacing free-form critique with structured verification questions \cite{Dhuliawala2023ChainofVerificationRH}. Evidence on SC's effectiveness is mixed, with documented failures on reasoning tasks and risks of negative revision \cite{Huang2023LargeLM,Kamoi2024WhenCL,Zhang2024UnderstandingTD,Yang2024ConfidenceVC}. Our work shifts focus from how to self-correct to when it works, identifying the task settings in which standard intrinsic revision reliably yields improvements. 

\section{Conclusion}
Motivated by prior work showing that intrinsic self-correction can be unreliable, we studied whether revision procedures can yield gains in targeted task settings. We evaluated self-correction on tasks involving explicit constraints, complex reasoning, and second-opinion revision over alternative candidates in word games. Across these settings, intrinsic self-correction improved performance over baseline prompting, with especially consistent gains on highly verifiable tasks. However, the benefits were not uniform: stronger models sometimes showed limited gains, and weaker models sometimes lacked the task capability to revise effectively. Our sub-task analysis further indicates that reasoning intensity and verifiability help explain where self-correction is most likely to help. Overall, these results support a task-sensitive view of intrinsic self-correction as an inference-time strategy whose usefulness depends on the role a second pass can play in a given task.

\section{Limitations}
This study focuses on a controlled evaluation of intrinsic self-correction (SC), and its scope is therefore necessarily limited. Our benchmark suite was selected to capture different revision-related properties, including verifiability, reasoning intensity, and second-opinion generation. However, it does not cover all possible use cases, particularly highly open-ended generation tasks such as long-form writing, summarization, or code synthesis, where evaluation criteria are less clearly defined.

To ensure reproducibility, our experiments use standard two-step and three-step prompting templates with greedy decoding. This design allows for a consistent comparison across models and tasks, but does not examine alternative inference-time strategies such as sampling, self-consistency, debate-style prompting, or longer iterative refinement procedures. These methods may lead to different outcomes and are left for future work.

Our sub-task analysis of BBEH uses strong LLMs as proxy judges to estimate task properties such as reasoning intensity and verifiability. While this enables scalable annotation, such judgments may reflect model-specific preferences. We therefore interpret these analyses as indicative rather than definitive.

Finally, LLM behavior continues to evolve as pre-training, post-training, and alignment methods improve. The patterns observed in this work, including cases of over-revision or limited self-verification, may change for future model generations. Continued evaluation will therefore be important for understanding how intrinsic SC develops over time.

\bibliography{references}

\newpage
\appendix

\section{Per-Task Analysis for BIG-Bench Extra Hard}
\label{sec:per_task_analysis}
\paragraph{Motivation}
We hypothesize that self-correction is more effective for tasks that require intensive reasoning or where solutions can be easily verified.

For example, in SAT, deriving a solution requires substantial reasoning to solve, yet its solutions can be verified in polynomial time. Consistent with this hypothesis, we observe that self-correction is highly effective for SAT.

To further study this question, and without arguing that this result is general, we decided to score each sub-task of BBEH \cite{kazemi2024boardgameqa, nie2024moca, kiciman2023causal, tyen2023llms, kazemi2023geomverse, sanchez2024linguini, hessel2022androids, zhang2024humor, yamada2023evaluating, fatemi2024test, white2024livebench, shah2024causal} based on how easy they are to \emph{verify}, and how much \emph{reasoning} they require. Then, we examine whether those \emph{reasoning and verifiability scores} are correlated with the gains from self-correction.

\paragraph{Setup}
We use LLM-judges to score each sub-task. Specifically, we prompted GPT-5.5 and Opus-4.6 to score each of those factors, after an extensive comparison to each of the other tasks. The judge is shown all task descriptions and 1 example for each. For the scored task, we present 5 random samples to reduce bias and calibrate the judge's score.

We describe the factors we study (verifiability and reasoning) minimally to prevent biases, although we acknowledge that the interpretation of verifiability or reasoning can be somewhat subjective. Additionally, we provide the judge with anchors (1, 50, 100), and briefly describe what each means (e.g., \emph{50 - Easier to verify than around half of the other tasks.}).

To ensure the (relative) reliability of the scores, we examined how correlated the two judges' scores are ($\rho=0.97$ for reasoning, $\rho=0.86$ for verifiability; both $p<0.001$). Then, we averaged their results and defined it as the final score per task.

\paragraph{Results}
For this discussion, we denote: \[\Delta = accuracy(SC)-accuracy(\text{Base})\]
and report the correlations (Spearman's $\rho$) of our reasoning and verifiability scores with the $\Delta$ of each benchmarked model. Since we study self-correction in a general sense, we are especially interested in the average $\Delta$ across the models. Note that the aggregate tends to correlate more strongly with our scores than the individual models do, which is a well-known effect \cite{Robinson1950EcologicalCA}. We observe that the average $\Delta$ across models is significantly correlated with both verifiability and reasoning. We also observe that the correlation is significant for about half of the models individually (see table \ref{tab:per_model_corr}). 

Additionally, we provide our reasoning and verifiability judge-scores, paired with $\Delta$, in Table \ref{tab:sc_gain_AVERAGE}. Notably, models tend to obtain a negative to negligible accuracy gain when self-correcting on tasks that are not verifiable and also require little reasoning. For all other tasks, it is difficult to identify a clear trend because $\Delta$ is influenced by many factors.

\paragraph{Confound Factors} One of our main concerns was that our reasoning and verifiability scores are correlated with baseline accuracy, hence serving as a proxy for a correlation between $\Delta$ and baseline performance. Within our sub-task sample, baseline accuracy shows little correlation with $\Delta$ ($\rho = 0.05$, $p = 0.83$), and moderate to weak correlation with reasoning scores ($\rho = -0.32$, $p = 0.14$) and verifiability scores ($\rho = 0.13$, $p = 0.56$). Additionally, we do not observe that our reasoning and verifiability scores are \emph{strongly} correlated ($\rho=0.19$, $p=0.39$).

\begin{table}[ht]
\centering
\begin{tabular}{lcc}
\toprule
Model & Reasoning & Verifiability \\
\midrule
Claude Haiku 4.5 & $0.19$ & $0.68^{*}$ \\
DeepSeek v3.1 & $0.28$ & $0.47^{*}$ \\
Gemini 3 Pro & $0.43^{*}$ & $0.13$ \\
Gemini 3.1 Flash Lite & $0.52^{*}$ & $0.43^{*}$ \\
Mistral Large & $0.36$ & $0.26$ \\
\midrule
Average & $0.54^{*}$ & $0.65^{*}$ \\
\bottomrule
\end{tabular}
\caption{Spearman $\rho$ between rubric score and $\Delta$ per model. Starred values are significant ($p<.05$).}
\label{tab:per_model_corr}
\end{table}

\iflatexml
  % LaTeXML has no xfp/cellcolor binding: just print the number, no cell shading.
  \newcommand{\basescore}[1]{#1}
  \newcommand{\verifscore}[1]{#1}
\else
  \definecolor{datagreen}{RGB}{0, 255, 0}
  \definecolor{datared}{RGB}{255, 0, 0}
  \newcommand{\basescore}[1]{\cellcolor{datagreen!\fpeval{round(#1 / 97 * 100)}!white}#1}
  \newcommand{\verifscore}[1]{\cellcolor{datared!\fpeval{round(#1 / 83 * 100)}!white}#1}
\fi

\begin{table*}[ht]
\centering
\begin{tabular}{lccr}
\toprule
Task & Reasoning Score & Verifiability Score & SC Gain ($\Delta$) \\
\midrule
Zebra Puzzles & \basescore{97} & \verifscore{36} & +16.2\% \\
Web Of Lies & \basescore{89} & \verifscore{62} & +13.4\% \\
Time Arithmetic & \basescore{60} & \verifscore{66} & +12.7\% \\
Object Counting & \basescore{34} & \verifscore{70} & +10.1\% \\
Word Sorting & \basescore{46} & \verifscore{74} & +7.9\% \\
Spatial Reasoning & \basescore{92} & \verifscore{36} & +7.7\% \\
Multistep Arithmetic & \basescore{94} & \verifscore{50} & +6.1\% \\
Boardgame Qa & \basescore{87} & \verifscore{58} & +5.9\% \\
Boolean Expressions & \basescore{66} & \verifscore{67} & +5.2\% \\
Buggy Tables & \basescore{90} & \verifscore{54} & +5.2\% \\
Shuffled Objects & \basescore{78} & \verifscore{53} & +4.6\% \\
Hyperbaton & \basescore{74} & \verifscore{42} & +4.3\% \\
Temporal Sequence & \basescore{90} & \verifscore{42} & +4.0\% \\
Object Properties & \basescore{82} & \verifscore{48} & +3.2\% \\
Dyck Languages & \basescore{66} & \verifscore{83} & +3.1\% \\
Linguini & \basescore{82} & \verifscore{26} & +1.9\% \\
Causal Understanding & \basescore{32} & \verifscore{23} & +1.1\% \\
Disambiguation Qa & \basescore{31} & \verifscore{54} & +1.0\% \\
Nycc & \basescore{18} & \verifscore{4} & +0.3\% \\
Geometric Shapes & \basescore{90} & \verifscore{32} & -2.1\% \\
Sportqa & \basescore{33} & \verifscore{17} & -2.5\% \\
Sarc Triples & \basescore{22} & \verifscore{18} & -3.4\% \\
Movie Recommendation & \basescore{12} & \verifscore{8} & -4.3\% \\
\bottomrule
\end{tabular}
\caption{BBEH per-task self-correction results (averaged across all models) sorted by accuracy change ($\Delta$). Positive $\Delta$ indicates that self-correction improved performance, negative $\Delta$ indicates degradation. Higher verifiability/reasoning scores correspond to higher color saturation.}
\label{tab:sc_gain_AVERAGE}
\end{table*}

\section{Extended Evaluation Details}
\label{sec:extended_eval_details}

\subsection{Model Selection and Evaluation Phasing}
Our experiments evaluate both open-weights and proprietary large language models. For open-weights architectures, we assess DeepSeek-Chat-v3.1 \cite{liu2024deepseek} and Mistral-Large-2512 \cite{mistral2024large}. Our proprietary suite includes Claude-4.5-Haiku \cite{anthropic2024claude} and Google's Gemini series \cite{team2023gemini}--specifically Gemini-3.1-Flash-Lite-Preview, Gemini-3-Pro-Preview, and Gemini-2.5-Pro. 

During the evaluation phase, API availability necessitated a planned transition within the Gemini suite. Initial experiments across the SAT, Big Bench Extra Hard (BBEH) Choice, BBEH Non-Choice, and Codenames benchmarks were conducted using the \texttt{gemini-3-pro-preview} endpoint. Following the mid-experiment deprecation of this preview variant, we transitioned to the generally available \texttt{gemini-2.5-pro} model to ensure continuity and reproducibility. Consequently, all results reported for the HLE Text Choice, Hangman, and Wordle benchmarks exclusively reflect the performance of this stable fallback model, which retained the robust reasoning capabilities requisite for these tasks.

\subsection{Codenames Agent Configuration}
In the Codenames benchmark, the environment simulates a multi-agent setup comprising two distinct roles: the guess giver (spymaster) and the guesser. For the purpose of evaluating intrinsic self-correction in strategic planning, our self-correction protocol was exclusively applied to the guess giver. The guesser agent operated in a standard, zero-shot generation mode without iterative refinement, ensuring that any performance variations were strictly attributable to the spymaster's revised clues.

\subsection{Handling Missing Responses}
To maintain the integrity of our comparative analysis, we applied a strict exclusion criterion during the evaluation phase. When calculating accuracy and task-specific scores, we omitted any instance where the LLM API failed to return a response--such as when the model stalled, timed out, or triggered a refusal. By skipping these empty generations, we ensure that the reported metrics reflect a direct comparison of reasoning trajectories rather than artifacts of API failures or abstain rates.

\subsection{Dataset Sample Sizes}
For each benchmark, we evaluate over a substantial number of samples to ensure reliable estimates. The precise number of task instances evaluated per dataset is outlined below:
\begin{itemize}
    \item \textbf{SAT:} 234 samples
    \item \textbf{BBEH (Non-Choice):} \textasciitilde 500 samples
    \item \textbf{BBEH (Choice):} 400 samples
    \item \textbf{HLE (Text Choice):} 200 samples
    \item \textbf{Wordle:} 200 games
    \item \textbf{Hangman:} 240 games
    \item \textbf{Codenames:} 400 games
\end{itemize}

\subsection{Detailed Revision Rates}
Table~\ref{tab:revision_rate} details the revision rates across all evaluated models and games. These empirical results strongly support our claim regarding the behavior of Claude Haiku 4.5 in Codenames (see Section \ref{sec:results}). Notably, Claude exhibits an exceptionally high revision rate of 98.7\%, meaning it alters its baseline answer in nearly every instance during the self-correction phase. This rate is not only objectively high but also significantly outpaces all other models on the same benchmark and Claude's own revision rate on other games.

\begin{table*}[t]
\centering
\resizebox{\linewidth}{!}{
\begin{tabular}{l c c c c c}
\toprule
Revision Rate (\%) & Mistral Large 2512 & Claude Haiku 4.5 & DeepSeek V3.1 & Gemini 3.1 Flash Lite & Gemini Pro \\
\midrule
Wordle & 0.0 & 30.3 & 3.2 & 45.4 & 1.3 \\
Hangman & 3.7 & 2.1 & 8.8 & 13.9 & 3.2 \\
Codenames & 47.8 & \textbf{98.7} & 77.2 & 85.0 & 38.4 \\
\bottomrule
\end{tabular}
}
\caption{Revision rates (\%) across models and tasks. The rate denotes the percentage of instances where the self-correction agent selected a different final answer compared to the baseline agent. For multi-turn games (Wordle and Hangman), results are averaged across all turns.}
\label{tab:revision_rate}
\end{table*}

\section{Prompt Templates}
\label{sec:prompt_templates}

Although all experiments follow the same intrinsic self-correction principle, we do not use a single universal prompt across benchmarks. Instead, each of the six benchmarks uses a task-specific prompt that preserves the same experimental structure while adapting the instructions, answer format, and constraints to the benchmark. This design avoids giving models irrelevant instructions and ensures that the base and self-correction stages are evaluated under the same task requirements.

Following recent concerns that apparent self-correction gains can be driven by unfair prompting choices rather than genuine revision ability \citep{Kamoi2024WhenCL}, we explicitly checked prompt fairness during prompt design. In particular, we tested that the self-correction prompt did not introduce new task information, hidden labels, oracle feedback, or additional constraints unavailable to the base prompt. The revision stage only receives the original task input and the model's own previous response, and is instructed to reconsider whether the answer should be kept or changed. This prevents the SC condition from benefiting from external feedback while still allowing the model to use an additional intrinsic reasoning pass.

\subsection{Direct-Refine Protocol}
For SAT, Wordle, Hangman, and Codenames benchmarks, we use a two-step protocol. The base prompt presents the benchmark-specific task description and required output format. The self-correction prompt then asks the same model to review the original input and its own previous answer, check the relevant task constraints, consider whether an alternative answer is preferable, and output a final revised answer in the same format.

\vspace{0.5em}
\noindent \textbf{Step 1: Base Generation} \\
\textit{Input:} \texttt{[Benchmark-specific task description, instance, constraints, and output format]} \\
\textit{Output:} \texttt{[Initial Answer]}

\vspace{0.5em}
\noindent \textbf{Step 2: Self-Correction} \\
\textit{Input:} \texttt{[Original task input + model's initial answer + benchmark-specific instruction to verify constraints and decide whether to keep or revise the answer]} \\
\textit{Output:} \texttt{[Final Revised Answer in the same benchmark-specific format]}

\subsection{Direct-Critic-Refine Protocol}
For BBEH and HLE benchmarks, we use a three-step protocol because these benchmarks require longer reasoning trajectories and benefit from separating critique from final revision. The critique prompt is still intrinsic: it asks the model to inspect its own reasoning for possible mistakes, unsupported assumptions, or inconsistencies, but does not provide correctness labels or external evidence. The revision prompt then asks the model to answer the original question again using only the generated critique and the original input.

\vspace{0.5em}
\noindent \textbf{Step 1: Base Generation} \\
\textit{Input:} \texttt{[Benchmark-specific question, answer choices if applicable, and output format]} \\
\textit{Output:} \texttt{[Initial Answer]}

\vspace{0.5em}
\noindent \textbf{Step 2: Intrinsic Critique} \\
\textit{Input:} \texttt{[Original task input + model's initial answer + instruction to identify possible reasoning errors or inconsistencies]} \\
\textit{Output:} \texttt{[Self-Generated Feedback]}

\vspace{0.5em}
\noindent \textbf{Step 3: Self-Correction} \\
\textit{Input:} \texttt{[Original task input + initial answer + self-generated feedback + instruction to provide a final answer in the benchmark-specific format]} \\
\textit{Output:} \texttt{[Final Revised Answer]}

\subsection{SAT Prompts}
\paragraph{Direct Agent Prompt}
\begin{promptbox}
[System Prompt]
You are an expert Boolean Satisfiability (SAT) solver. You will be given CNF formulas 
that are SATISFIABLE and must find a valid variable assignment.

Analyze the formula and provide an assignment that satisfies ALL clauses.

Format your final answer as:
ASSIGNMENT: x1=True, x2=False, x3=True, x4=False, ...

[Task Prompt]
Find a satisfying assignment for this Boolean Satisfiability (SAT) problem:
Variables: {task.num_vars} (x1, x2, ..., x{task.num_vars})
Clauses: {task.num_clauses}

CNF Formula (Conjunctive Normal Form):
  Clause {i}: ({' v '.join(literals)})

Find values for all variables (True or False) that satisfy ALL clauses.
Remember: A clause is satisfied if at least ONE of its literals is true.

[Trailing User Prompt]
(Remember to provide: ASSIGNMENT: x1=True/False, x2=True/False, ... for all variables)
\end{promptbox}
\paragraph{Critic+Refinement Agent Prompt}
\begin{promptbox}
You are a strict, skeptical SAT solution verifier.

---------------------------
PROBLEM
---------------------------
{question}

---------------------------
PROPOSED SOLUTION
---------------------------
{direct_answer}

---------------------------
YOUR TASK
---------------------------
Verify if the proposed assignment actually satisfies ALL clauses.

VERIFICATION PROCESS:
- Check each clause one by one
- A clause (a v b v c) is satisfied if ANY literal is true
- Literal x_i is true if variable x_i = True
- Literal \neg x_i is true if variable x_i = False
- ALL clauses must be satisfied

CRITICAL CHECKS:
1. COMPLETENESS: Are all variables assigned?
2. VALIDITY: Does the assignment satisfy EVERY clause?
3. ERRORS: Are there any mistakes in the reasoning?

---------------------------
OUTPUT FORMAT
---------------------------
First, verify each clause step by step.
Then:
- If the solution is CORRECT, output: "VERIFIED CORRECT"
- If there are errors, explain them and provide a corrected assignment

Your final answer must include:
ASSIGNMENT: x1=True/False, x2=True/False, ... for ALL variables
\end{promptbox}

\subsection{BBEH Non Choice Prompts}
\paragraph{Direct Agent Prompt}
\begin{promptbox}
[System Prompt]
You are a precise short-answer engine. You must answer with the exact 
short string/number requested.
Format your final answer exactly as: Answer:(<answer>)

[Trailing User Prompt]
(Remember to end your response with exactly: Answer:(<answer>) )
\end{promptbox}
\paragraph{Critic Agent Prompt}
\begin{promptbox}
You are a strict, skeptical, and meticulous critic.

---------------------------
QUESTION
---------------------------
{question}

---------------------------
PROPOSED ANSWER
---------------------------
{direct_answer}

---------------------------
YOUR TASK
---------------------------
Evaluate the PROPOSED ANSWER solely on its logical merit and factual accuracy.

CRITICAL INSTRUCTIONS:
1. NEUTRAL STANCE: Do not assume the answer is correct, and do not assume it is wrong. Judge it purely on the evidence provided.
2. CONSISTENCY CHECK: Does the reasoning provided logically lead to the selected conclusion? Are there contradictions?
3. COMPLETENESS: Did the answer address the *exact* question asked (including specific constraints or units)?
4. EVIDENCE BASED: Only flag an error if you can identify a specific factual mistake or logical fallacy. Do not critique stylistic choices.

---------------------------
OUTPUT
---------------------------
- If the reasoning is sound and the answer is correct, output: "No errors found."
- If there is a demonstrable error in facts or logic, list it clearly.

Do NOT generate a new answer.
Do NOT output the corrected letter choice yet.
\end{promptbox}
\paragraph{Refinement Agent Prompt}
\begin{promptbox}
You are a precise short-answer engine.

---------------------------
QUESTION
---------------------------
{question}

---------------------------
PREVIOUS ATTEMPT
---------------------------
{direct_answer}

---------------------------
CRITIQUE OF PREVIOUS ATTEMPT
---------------------------
{critic_answer}

---------------------------
YOUR TASK
---------------------------
Using the insights from the CRITIQUE, write a new, correct solution to the QUESTION.
1. If the critique found errors, fix them in your reasoning.
2. If the critique found no errors, maintain the original logic.

CONSTRAINTS:
- Do NOT apologize or mention that you are correcting a mistake.
- Do NOT output the critique text again.
- Go straight to the reasoning and the final answer.

---------------------------
CRITICAL OUTPUT FORMAT
---------------------------
You may think step by step, but your response MUST end with EXACTLY:

Answer:(<answer>)

Where <answer> is the exact short string/number requested.
\end{promptbox}

\subsection{BBEH Choice Prompts}
\paragraph{Direct Agent Prompt}
\begin{promptbox}
You are a precise multiple-choice answering engine. You must answer with ONE letter option.
Format your final answer exactly as: Answer:(X.)
{question}

(Remember to end your response with exactly: Answer:(X.) )
\end{promptbox}
\paragraph{Direct Agent Prompt}
\begin{promptbox}
You are a precise multiple-choice answering engine. You must answer with ONE letter option.
Format your final answer exactly as: Answer:(X.)
{question}

(Remember to end your response with exactly: Answer:(X.) )
\end{promptbox}
\paragraph{Critic Agent Prompt}
\begin{promptbox}
You are a strict, skeptical, and meticulous critic.

---------------------------
QUESTION
---------------------------
{question}

---------------------------
PROPOSED ANSWER
---------------------------
{direct_answer}

---------------------------
YOUR TASK
---------------------------
Evaluate the PROPOSED ANSWER solely on its logical merit and factual accuracy.

CRITICAL INSTRUCTIONS:
1. NEUTRAL STANCE: Do not assume the answer is correct, and do not assume it is wrong. Judge it purely on the evidence provided.
2. CONSISTENCY CHECK: Does the reasoning provided logically lead to the selected conclusion? Are there contradictions?
3. COMPLETENESS: Did the answer address the *exact* question asked (including specific constraints or units)?
4. EVIDENCE BASED: Only flag an error if you can identify a specific factual mistake or logical fallacy. Do not critique stylistic choices.

---------------------------
OUTPUT
---------------------------
- If the reasoning is sound and the answer is correct, output: "No errors found."
- If there is a demonstrable error in facts or logic, list it clearly.

Do NOT generate a new answer.
Do NOT output the corrected letter choice yet.
\end{promptbox}
\paragraph{Refinement Agent Prompt}
\begin{promptbox}
You are a precise answering engine.

---------------------------
QUESTION
---------------------------
{question}

---------------------------
PREVIOUS ATTEMPT
---------------------------
{direct_answer}

---------------------------
CRITIQUE OF PREVIOUS ATTEMPT
---------------------------
{critic_answer}

---------------------------
YOUR TASK
---------------------------
Using the insights from the CRITIQUE, write a new, correct solution to the QUESTION.
1. If the critique found errors, fix them in your reasoning.
2. If the critique found no errors, maintain the original logic.

CONSTRAINTS:
- Do NOT apologize or mention that you are correcting a mistake.
- Do NOT output the critique text again.
- Go straight to the reasoning and the final answer.

---------------------------
CRITICAL OUTPUT FORMAT
---------------------------
You may think step by step, but your response MUST end with EXACTLY:

Answer:(X.)

Where X is just the letter of your final choice (A, B, C, D, E, or F).
\end{promptbox}

\subsection{HLE Choice Prompts}
\paragraph{Direct Agent Prompt}
\begin{promptbox}
You are an expert solving an advanced academic problem. Read the problem carefully, think step-by-step, and provide your detailed reasoning. 

You must conclude your response by stating your final answer on a new, separate line using exactly the following format:
Answer: (X.)

Problem:
{question}
\end{promptbox}
\paragraph{Critic Agent Prompt}
\begin{promptbox}
You are an expert reviewer evaluating a proposed solution to an advanced academic problem.
Carefully review the problem and the proposed solution. Identify any logical flaws, calculation errors, empirical inaccuracies, or incorrect assumptions in the reasoning. 

Provide only your critique of the proposed solution.

Problem:
{question}

Proposed Solution:
{direct_answer}

Critique:
\end{promptbox}
\paragraph{Refinement Agent Prompt}
\begin{promptbox}
You are an expert solving an advanced academic problem. You are provided with a problem, a previous attempt at a solution, and a critique of that attempt. 

Carefully consider the critique. Correct any identified mistakes and provide a revised, step-by-step reasoning. 

You must conclude your response by stating your final answer on a new, separate line using exactly the following format:
Answer: (X.)

Problem:
{question}

Previous Attempt:
{direct_answer}

Critique:
{critic_answer}

Revised Solution:
\end{promptbox}

\subsection{Wordle Prompts}
\paragraph{Direct Agent Prompt}
\begin{promptbox}
[System Prompt]
You are an expert Wordle player. Guess the secret 5-letter word within 6 attempts.

After each guess you receive letter-by-letter feedback:
  G (Green)  -- correct letter in the correct position
  Y (Yellow) -- letter is in the word but in the wrong position
  X (Grey)   -- letter is not in the word at all

Strategy tips:
- Green letters must stay in the same position in future guesses.
- Yellow letters must appear somewhere else in future guesses.
- Grey (X) letters should not be reused.

Format your guess as [word]  (e.g., [crane])
One guess per turn.

[User Prompts]
A {word_length}-letter word has been chosen. 
You have {max_guesses} attempts.
Guess #1:

Feedback: {word.upper()} -> {fb_str}

Guess #{next_turn}:
\end{promptbox}
\paragraph{Critic+Refinement Agent Prompt}
\begin{promptbox}
[System Prompt]
You are an expert Wordle player. Guess the secret 5-letter word within 6 attempts.

After each guess you receive letter-by-letter feedback:
  G (Green)  -- correct letter in the correct position
  Y (Yellow) -- letter is in the word but in the wrong position
  X (Grey)   -- letter is not in the word at all

Strategy tips:
- Green letters must stay in the same position in future guesses.
- Yellow letters must appear somewhere else in future guesses.
- Grey (X) letters should not be reused.

Another player has suggested a guess. Consider whether it follows the rules and 
whether you can think of a better option. You can either go with it (output ACCEPT) 
or play a different word instead (output [word]).
End your response with either ACCEPT or [word].

[User Prompt]
Word length: {word_length} letters  |  Turn: #{turn_num} of {max_guesses}  |  Remaining guesses: {remaining}

Game history:
{history_text}

Proposed next guess: [{proposed_guess.upper()}]

Review the proposed guess and respond with either ACCEPT or [better_word].
\end{promptbox}

\subsection{Hangman Prompts}
\paragraph{Direct Agent Prompt}
\begin{promptbox}
[System Prompt]
You are playing Hangman. Guess the hidden word one letter at a time, or guess the full word when confident.

Rules:
- Guess a single letter: [A]
- Guess the full word: [HELLO]
- A correct letter is revealed in all its positions.
- A wrong letter costs you one of your 6 allowed mistakes.
- A wrong full-word guess does NOT cost a mistake -- but only guess the word when you are fairly sure.
- Do not guess a letter you have already tried.

Start with common letters (E, T, A, O, I, N, S, R) and use the revealed pattern to narrow down the word.

[User Prompt]
{board_str}

Guess #{turn_num}:
\end{promptbox}
\paragraph{Critic+Refinement Agent Prompt}
\begin{promptbox}
[System Prompt]
You are playing Hangman. Guess the hidden word one letter at a time, or guess the full word when confident.

Rules:
- Guess a single letter: [A]
- Guess the full word: [HELLO]
- A correct letter is revealed in all its positions.
- A wrong letter costs you one of your 6 allowed mistakes.
- A wrong full-word guess does NOT cost a mistake -- but only guess the word when you are fairly sure.
- Do not guess a letter you have already tried.

Another player has suggested a guess. Consider whether it follows the rules and 
whether you can think of a better option. You can either go with it (output ACCEPT) 
or play a different guess instead (output [letter] or [word]).
End your response with either ACCEPT or your guess in brackets.

[User Prompt]
{board_str}

Proposed guess: {proposed_display}

Review this guess and respond with either ACCEPT or your own guess in brackets.
\end{promptbox}

\subsection{Codenames Prompts}
\paragraph{Direct Agent Prompt}
\begin{promptbox}
You are the Clue Giver in a word association game.

GAME SETUP:
- There are 6 words on the board. 3 are "special" (your targets) and 3 are "decoys".
- You must give ONE single-word clue that connects your 3 special words, without pointing to the decoys.
- The guesser will see all 6 words and your clue, then decide how many words to guess (1, 2, or 3).
- Scoring for the guesser: +1 per correct guess, -1 per wrong guess. So be clear!

All 6 words: {', '.join(all_words)}
Your 3 SPECIAL words (target): {', '.join(special_words)}
3 DECOY words (avoid): {', '.join(decoys)}

STRATEGY:
- Find a connection that fits all 3 special words but NOT the decoys.
- Your clue cannot be any of the 6 words on the board.
- Your clue must be a single English word.

OUTPUT FORMAT (strict):
REASONING: [brief explanation of the connection and why it avoids the decoys]
FINAL CLUE: word
\end{promptbox}
\paragraph{Critic+Refinement Agent Prompt}
\begin{promptbox}
You are the CRITIC for the Clue Giver in a word association game.

GAME SETUP:
- 6 words on the board: {', '.join(all_words)}
- 3 SPECIAL words (targets): {', '.join(special_words)}
- 3 DECOY words (avoid): {', '.join(decoys)}
- Scoring for the guesser: +1 per correct guess, -1 per wrong guess.

PROPOSED CLUE (from main agent):
"{answer_a['clue']}"

Previous reasoning:
{answer_a['raw']}

YOUR JOB:
Evaluate the proposed clue on two criteria:
1. COVERAGE: Does it clearly connect all 3 special words?
2. SAFETY: Could it lead the guesser to pick a decoy word instead?

If the clue is good -> ACCEPT it unchanged.
If the clue is risky or weak -> OVERRIDE with a better one.

OUTPUT FORMAT (strict):
ANALYSIS: [Your evaluation of coverage and safety]
DECISION: ACCEPT or OVERRIDE
FINAL CLUE: word
\end{promptbox}

\section{Example tasks of our custom benchmarks}
\paragraph{SAT example task}
\begin{promptbox}
Find a satisfying assignment for this Boolean Satisfiability (SAT) problem:
Variables: 19 (x1, x2, ..., x19)
Clauses: 40

CNF Formula (Conjunctive Normal Form):
  Clause 1: (x14 v x19)
  Clause 2: (x1 v x15 v \neg x19)
  Clause 3: (\neg x10 v x8 v x12)
  Clause 4: (\neg x16 v x3 v x6)
  Clause 5: (x17 v \neg x18)
  Clause 6: (x1 v \neg x15)
  Clause 7: (\neg x2 v x18 v \neg x6)
  Clause 8: (x18 v \neg x19 v x5 v \neg x14)
  Clause 9: (\neg x4 v \neg x3 v \neg x18 v x2)
  Clause 10: (x18 v \neg x6)
  Clause 11: (\neg x4 v \neg x9)
  Clause 12: (x18 v \neg x19 v x14 v x15)
  Clause 13: (\neg x10 v \neg x7 v x1)
  Clause 14: (\neg x10 v x4)
  Clause 15: (\neg x16 v x8 v x2)
  Clause 16: (\neg x6 v x12 v \neg x10)
  Clause 17: (\neg x8 v x19)
  Clause 18: (x11 v x7)
  Clause 19: (x9 v \neg x13)
  Clause 20: (x14 v x12 v \neg x6 v \neg x3)
  Clause 21: (\neg x3 v x10 v x18)
  Clause 22: (\neg x10 v x15 v x3 v x14)
  Clause 23: (x9 v x6 v x18 v x13)
  Clause 24: (\neg x19 v x18 v x3)
  Clause 25: (\neg x15 v \neg x17 v \neg x14)
  Clause 26: (\neg x4 v \neg x6 v \neg x14)
  Clause 27: (x12 v \neg x2 v \neg x6)
  Clause 28: (\neg x4 v x16)
  Clause 29: (x12 v \neg x15 v \neg x1 v \neg x6)
  Clause 30: (\neg x4 v \neg x7 v x8 v \neg x16)
  Clause 31: (\neg x10 v \neg x18 v \neg x9)
  Clause 32: (x13 v x10 v \neg x16)
  Clause 33: (\neg x8 v x15 v x18)
  Clause 34: (\neg x8 v \neg x14 v x7)
  Clause 35: (\neg x15 v x5 v x7 v \neg x18)
  Clause 36: (x14 v \neg x10 v x13 v \neg x12)
  Clause 37: (\neg x8 v \neg x15 v \neg x5)
  Clause 38: (x12 v \neg x13)
  Clause 39: (\neg x8 v \neg x19)
  Clause 40: (\neg x16 v x7 v \neg x11 v \neg x2)

Find values for all variables (True or False) that satisfy ALL clauses.
Remember: A clause is satisfied if at least ONE of its literals is true.

(Remember to provide: ASSIGNMENT: x1=True/False, x2=True/False, ... for all variables)
\end{promptbox}

\paragraph{Wordle example task}
\begin{promptbox}
*system message:*
You are an expert Wordle player. Guess the secret 5-letter word within 6 attempts.

After each guess you receive letter-by-letter feedback:
  G (Green)  - correct letter in the correct position
  Y (Yellow) - letter is in the word but in the wrong position
  X (Grey)   - letter is not in the word at all

Strategy tips:
- Green letters must stay in the same position in future guesses.
- Yellow letters must appear somewhere else in future guesses.
- Grey (X) letters should not be reused.

Format your guess as [word]  (e.g., [crane])
One guess per turn.

*First user message:*
A 5-letter word has been chosen. You have 6 attempts.
Guess #1:
\end{promptbox}

\paragraph{Hangman example task}
\begin{promptbox}
*system message:*
You are playing Hangman. Guess the hidden word one letter at a time, or guess the full word when confident.

Rules:
- Guess a single letter: [A]
- Guess the full word: [HELLO]
- A correct letter is revealed in all its positions.
- A wrong letter costs you one of your 6 allowed mistakes.
- A wrong full-word guess does NOT cost a mistake - but only guess the word when you are fairly sure.
- Do not guess a letter you have already tried.

Start with common letters (E, T, A, O, I, N, S, R) and use the revealed pattern to narrow down the word.

*First user message:*
Word: _ _ _ _ _ _
Guessed letters: None
Wrong guesses left: 6 / 6

Guess #1:

\end{promptbox}

\paragraph{Codenames example task}
\begin{promptbox}
*Agent A - Clue Giver*

You are the Clue Giver in a word association game.

GAME SETUP:
- There are 6 words on the board. 3 are "special" (your targets) and 3 are "decoys".
- You must give ONE single-word clue that connects your 3 special words, without pointing to the decoys.
- The guesser will see all 6 words and your clue, then decide how many words to guess (1, 2, or 3).
- Scoring for the guesser: +1 per correct guess, -1 per wrong guess. So be clear!

All 6 words: singer, long, tattoo, whiteboard, rope, sole
Your 3 SPECIAL words (target): singer, tattoo, long
3 DECOY words (avoid): whiteboard, rope, sole

STRATEGY:
- Find a connection that fits all 3 special words but NOT the decoys.
- Your clue cannot be any of the 6 words on the board.
- Your clue must be a single English word.

OUTPUT FORMAT (strict):
REASONING: [brief explanation of the connection and why it avoids the decoys]
FINAL CLUE: word

*Agent B - Guesser*
You are the Guesser in a word association game.

GAME SETUP:
- There are 6 words on the board. 3 are "special" and 3 are "decoys".
- The Clue Giver gave you the clue: "live"
- You must guess which words are special based on this clue.

The 6 words: singer, long, tattoo, whiteboard, rope, sole

SCORING:
- +1 point for each correct word you guess
- -1 point for each wrong word you guess
- You can guess 1, 2, or 3 words - guess fewer if you're unsure to avoid losing points.

STRATEGY:
- Rank all 6 words by how strongly they connect to the clue "live".
- Only include a word in your guess if you're reasonably confident it's special.
- If only 1 or 2 words feel clearly connected, just guess those.

OUTPUT FORMAT (strict):
ANALYSIS: [For each word, rate its connection to the clue: strong/medium/weak/none]
FINAL GUESS: word1, word2  (list only the words you choose to guess, separated by commas)


\end{promptbox}

\section{Judge Prompts}
\paragraph{Judge (verifiability)}
\begin{promptbox}
[System Prompt]
Rate how easy it is to verify whether a proposed answer is correct, given the
task input and the answer. Assume external information is never available (e.g., you can't search the web!).

Anchors (1, 50, and 100 are anchored; all other integers are valid
intermediate positions):
  1    Impossible or very difficult to verify.
  50   Easier to verify than around half of the other tasks.
  100  Easier to verify than all other tasks.

Ideally, the easiest task to verify out of this set will score 100, and tasks which are harder to verify will score lower.

Steps:
  1. Imagine a candidate answer in hand.
  2. Compare this to 22 other tasks one by one (state that you do "Comparison i, comparing task $task with $other_task"), how easy is it to verify the answer?
  3. Place the task on the 1->100 continuum (exactly based on your comparisons and the anchors defined above).

Respond with STRICT JSON only, matching this schema:
{
  "reasoning": "<walk through each Step; for the comparison step, compare against every other task explicitly>",
  "score": <integer 1-100>,
  "justification": "<one short sentence restating the score>"
}
The "reasoning" field MUST appear before "score". Do not include any text outside the JSON object.

[User Prompt]
=== All benchmark tasks (for calibration --- read these to understand the full range before scoring) ===

--- {task 1 name} ---
{task 1 README}

--- Example 1 ---
INPUT: {task 1 example input}
TARGET: {task 1 example target}

...

--- {task 23 name} ---
{task 23 README}

--- Example 1 ---
INPUT: {task 23 example input}
TARGET: {task 23 example target}

========================
Now score the following task:

Task name: {task name}

=== README.md ===
{task README / description}

=== 5 sampled examples (input + target) ===

--- Example 1 ---
INPUT:
{example 1 input}
TARGET: {example 1 target}

...

--- Example 5 ---
INPUT:
{example 5 input}
TARGET: {example 5 target}

Return the JSON now.
\end{promptbox}

\paragraph{Judge (reasoning)}
\begin{promptbox}
[System Prompt]
Rate how reasoning intensive a task is.

Anchors (1, 50, and 100 are anchored; all other integers are valid
intermediate positions):
  1    No reasoning required.
  50   Requires more reasoning than around half of the other tasks.
  100  Requires more reasoning than all other tasks.

Ideally, the most reasoning intensive task out of this set will score 100, and tasks which require less reasoning will score lower.

Steps:
  1. Consider what it takes to solve this task.
  2. Compare this to 22 other tasks one by one (state that you do "Comparison i, comparing task $task with $other_task"), how much reasoning is needed to produce the answer?
  3. Place the task on the 1->100 continuum (exactly based on your comparisons and the anchors defined above).

Respond with STRICT JSON only, matching this schema:
{
  "reasoning": "<walk through each Step; for the comparison step, compare against every other task explicitly>",
  "score": <integer 1-100>,
  "justification": "<one short sentence restating the score>"
}
The "reasoning" field MUST appear before "score". Do not include any text outside the JSON object.

[User Prompt]
=== All benchmark tasks (for calibration --- read these to understand the full range before scoring) ===

--- {task 1 name} ---
{task 1 README}

--- Example 1 ---
INPUT: {task 1 example input}
TARGET: {task 1 example target}

...

--- {task 23 name} ---
{task 23 README}

--- Example 1 ---
INPUT: {task 23 example input}
TARGET: {task 23 example target}

========================
Now score the following task:

Task name: {task name}

=== README.md ===
{task README / description}

=== 5 sampled examples (input + target) ===

--- Example 1 ---
INPUT:
{example 1 input}
TARGET: {example 1 target}

...

--- Example 5 ---
INPUT:
{example 5 input}
TARGET: {example 5 target}

Return the JSON now.
\end{promptbox}

\end{document}